\documentclass[11pt]{article}
\usepackage{fullpage}
\usepackage{times}
\usepackage{amsmath,amsthm,amsfonts}
\usepackage[usenames,dvipsnames]{color}
\usepackage{graphicx}
\usepackage{mathrsfs}
\usepackage{algorithm,algorithmic}
\usepackage{url}

\newcommand{\resolve}{{\sc resolwe}}
\newcommand{\baseline}{{\sc skipSelection}}
\newcommand{\set}[1]{\ensuremath{\mathcal{#1}}}

\begin{document}
\sloppy
\title{Structure Selection from \\ Streaming Relational Data}
\author{Lilyana Mihalkova \& Walaa Eldin Moustafa \\ University of Maryland College Park}
\date{May 1, 2011}
\maketitle
\begin{abstract}
Statistical relational learning techniques have been successfully applied in a wide range of relational domains. In most of these applications, the human designers capitalized on their background knowledge by following a trial-and-error trajectory, where relational features are manually defined by a human engineer, parameters are learned for those features on the training data, the resulting model is validated, and the cycle repeats as the engineer adjusts the set of features. This paper seeks to streamline application development in large relational domains by introducing a light-weight approach that efficiently evaluates relational features on pieces of the relational graph that are streamed to it one at a time. We evaluate our approach on two social media tasks and demonstrate that it leads to more accurate models that are learned faster.  
\end{abstract}
\section{Introduction}
Many machine learning applications involve inherently multi-relational domains in which entities of heterogeneous types engage in a variety of relations. The statistical relational learning (SRL) \cite{getoor:book07} community has introduced representations that provide principled support for learning and reasoning in such multi-relational data. In a nutshell, SRL models use an expressive relational language, such as first-order logic or SQL, to define relational features capable of capturing salient aspects of the structure of the domain. These relational features come with a parameterization, such that, once instantiated, they define a graphical model over which probabilistic inference can be performed. SRL techniques have been successfully applied in domains as diverse as biology, natural language processing, ontology alignment, social networks, and the web.

The basic design cycle of many such applications follows a trial-and-error trajectory, where relational features are manually defined by a human engineer, parameters are learned for those features on the training data, the resulting model is validated, and the cycle repeats as the human engineer adjusts the set of features. For example, this is the strategy recommended in the Markov logic network tutorial~\cite{domingos:icml07tutorial}\footnote{Available under ``Tutorial'' on \url{http://alchemy.cs.washington.edu/}}, slide 108, that comes with the widely used Alchemy software \cite{kok:tr05}.  Typically, as a result, a relatively small number of relational features are identified and used. This approach is appealing because background knowledge about the domain can be easily encoded in the intuitive relational language used in the SRL model at hand, and the relative strengths of these features can be learned through the parameterization.

An alternative to this design cycle is to use structure learning, where both the relational features and their accompanying parameterization are induced automatically from data. The state-of-the-art in structure learning has been advanced significantly in recent years \cite{getoor:jmlr02,kok:icml05,mihalkova:icml07,biba:ilp08,huynh:icml08,kok:icml09,kok:icml10,khosravi:aaai10}. This work has resulted in highly innovative approaches to identifying potentially promising structure candidates and efficiently navigating through the large and complex search space for structures. So far, less emphasis has been placed on how, once identified, structure candidates are evaluated. Existing techniques employ different flavors of a batch evaluation procedure, where candidate structures are scored on the available data and retained in the model if they improve the score. Crucially, all existing techniques presume that all of the data, or, at least joins of pairs of relational tables \cite{khosravi:aaai10}, can be stored in memory for the purposes of candidate structure evaluation.

While the importance of existing structure learning approaches is undisputed, their focus on identifying promising candidate structures, coupled with the assumption of batch scoring from in-memory data, is not a good match to all application scenarios. In particular, the designers of many SRL applications skipped structure learning entirely and instead preferred the trial-and-error design cycle outlined above. There are at least two plausible reasons underlying their chosen approach. First, the applications frequently involved data sets that were simply too large to allow for batch scoring of candidate structures. To overcome this,  parameters over  hand-coded relational features have been trained in parallel \cite{poon:emnlp08}, or on a stream of examples \cite{mihalkova:ecml09}.  Beyond the problem of storing large amounts of data in memory, in some domains, the data may actually arrive as a stream and not be available all at once.

Second, the designers of many applications often already have intuitions about good relational features, either from existing domain knowledge, such as in natural language or social network tasks, or from having worked with the ``raw'' data and developed a representation for it that is already biased by what they intuitively perceive as important aspects to capture. In such cases, where domain knowledge allows one to identify what variables are likely to influence each other, the design bottleneck is in pinning down the exact formulation of the influences among variables and eliminating features based on spurious intuitions. In other words, the problem of evaluating candidate features efficiently is at least as important as suggesting them.

This paper seeks to streamline application development in relational domains by introducing an approach that efficiently evaluates relational features on pieces of the relational graph that are streamed to it one at a time. We call our approach \resolve, for \underline{Re}lational \underline{S}tructure \underline{S}election from \underline{O}nline \underline{L}ight-\underline{W}eight \underline{E}valuation. \resolve\ is agnostic to where the relational features it evaluates come from. They could be discovered locally from each piece of the data that is being streamed using an existing structure discovery approach, or provided by an application developer. In this paper we take a semi-automated approach where the human designer specifies a declarative bias, e.g., \cite{deraedt:mlj97}, by providing a grammar for the relational features, from which all possible features are generated. This corresponds to the scenario outlined above where general intuitions about the domain are available, but determining the precise formulation of features requires efficiently evaluating different versions of the model on a large relational data set. 

We implement our approach in the framework of Markov logic networks (MLNs) \cite{richardson:mlj06}. This choice is motivated by the fact that MLNs have been widely used for relational application development. We start by providing necessary background in Section~\ref{sect:background}. \resolve\ is described in Section~\ref{sect:resolve}. In Section~\ref{sect:experiments}, we flesh out our proposed technique by using it to develop two applications in relational social media domains. Our results indicate that \resolve\ leads to significantly faster and more accurate learning.  We conclude with a discussion of related and future work.
\section{Background, Notation, Assumptions}
\label{sect:background}
{\bf First-order logic terminology.} In first-order logic, relations are represented as {\it predicates,} such as {\tt articleEdit(article, editor)}, which are Boolean functions with typed arguments. Assuming that the domain contains no functions, a {\it term} is defined as a variable or a constant. An {\it atom} is a predicate applied to terms, such as {\tt articleEdit(A, E)}. A positive/negative {\it literal} is a non-negated/negated atom. A literal is {\it grounded} if all of its arguments are constants, or actual entities from the domain; conversely, a literal is {\it ungrounded} if all of its arguments are variables. A formula consists of literals connected by conjunction and disjunction. Formulas whose literals contain only constants are grounded, whereas formulas whose literals contain only variables are ungrounded. We will refer to grounded formulas as {\it groundings}.

{\bf Learning Setting: }  We assume fully observable training data that consists of a large relational graph $\mathcal{G}$ whose hyperedges and nodes can correspond to different relations and entity types respectively. In practice, $\mathcal{G}$ is typically too large to fit in memory all at once, and/or parts of it arrive as learning progresses. Thus, we address a scenario where subgraphs of \set{G} arrive in a data stream \set{S}.  Second, we assume a discriminative setting where one or more relations in a set $\set{P}_T$ are designated as target predicates whose values are to be predicted at test time, and the remaining  relations $\set{P}_E$ are the evidence predicates whose values are observed at test time. 

{\bf Markov Logic Networks.} A Markov logic network (MLN) \cite{richardson:mlj06} consists of a set of first-order logic formulae $\mathcal{F}$, each of which has an associated weight. MLNs can be viewed as relational analogs to Markov networks, in which the potential functions over cliques are defined by the groundings of the formulae in $\mathcal{F}$.  The role of first-order logic, therefore, is to provide a highly expressive language for specifying general relational features.

Grounding the first-order logic formulae with a given set of entities  results in a Markov network. In particular, an MLN computes the conditional joint probability of a set of  predicate groundings $\mathbf{X}$ of the target predicates $\set{P}_T$, given truth values for a set of evidence predicate groundings $\mathbf{Y}$ as follows:
\begin{equation}
P(\mathbf{X} = \mathbf{x}| \mathbf{Y} = \mathbf{y}) = \frac{\exp(\sum_{f_i \in \mathcal{F}} w_i n_i(\mathbf{x, y}))} {\sum_{\mathbf{x'}}\exp(\sum_{f_i \in \mathcal{F}} w_i n_i(\mathbf{x', y}))}
\label{eqn:condProb}
\end{equation}
Above, $\mathbf{X}$ and $\mathbf{Y}$ are the sets of all target and evidence predicate groundings, respectively; $\mathbf{x}$ and $\mathbf{y}$ are the sets of corresponding truth assignments; $w_i$ is the weight associated with formula $f_i$; $n_i(\mathbf{x, y})$ is the number of {\tt true} groundings of formula $f_i$ on truth assignment $\mathbf{x, y}$; and the denominator computes the normalizing partition function $Z$. 

For ease of exposition, in this work we assume that all of the formulas in the MLN are conjunctions. This is not a restrictive assumption for the following reason. The most mature implementation of MLNs, Alchemy \cite{kok:tr05}, handles arbitrary formulas by converting them into conjunctive normal form, as a conjunction of disjunctions. Each disjunction produced in this way is then treated as a separate formula in the MLN, i.e., by viewing the MLN as a soft conjunction of disjunctions. Each disjunction in the MLN can then be converted to a conjunction by negating it, and also negating its weight. 
\section{The \resolve\ Algorithm}
\label{sect:resolve}
Learning on the stream \set{S} proceeds in a three-stage process:
\begin{enumerate}
\item The first $k_1$ sub-graphs that arrive are used to generate a set of relational features (in our case, first-order logic MLN formulae) \set{F}.
\item \resolve\ uses the next $k_2$ sub-graphs from \set{S} to evaluate the formulae in \set{F}, outputting a subset $\set{F}^* \subset \set{F}$ consisting of the formulae that, together, give good predictive accuracy for the groundings of $\set{P}_T$ given as observations the groundings of $\set{P}_E$.
\item The remainder of \set{S} is used to train parameters on the formulae in $\set{F}^*$.
\end{enumerate}

In this paper, the set of formulas \set{F} in step 1 above is generated without requiring any data from the stream (i.e., $k_1 = 0$) by using a declarative bias (in the spirit of \cite{deraedt:mlj97}) to specify templates, from which all possible formulas that comply with the bias are formed. Details on the templates used in each of our domains are provided in Section~\ref{sect:experiments}. Our goal in using this semi-automatic procedure to generate \set{F} was to take advantage of available background knowledge, while using systematic rule generation to make sure we do not inadvertently place the baseline to which we compare at a disadvantage
\subsection{Criteria for Effective Formulae}
 The goal of \resolve\ is to provide a light-weight formula evaluation strategy that can be carried out by considering sub-graphs of the training data graph $\mathcal{G}$ that arrive in a stream one at a time. The key is to develop criteria for what makes an effective formula and ensure that these criteria can be evaluated efficiently on \set{S}. For this purpose, it is useful to rewrite each formula $F$ as $E \wedge Q$, where $E$ is the evidence sub-formula and consists of all literals of $F$ with predicates in $\set{P}_E$, $Q$ is the query sub-formula and consists of all literals of $F$ with predicates  in $\set{P}_T$.\footnote{As described in Section~\ref{sect:background}, $F$ is assumed to be a conjunction.} We can view the roles of $E$ and $Q$ in $F$ as {\it selector} and {\it enforcer} respectively: groundings of $F$ in which the corresponding grounding of $E$ is satisfied are ``selected'' and a particular pattern, or configuration of truth values, specified by $Q$ is ``enforced'' over the truth values of the corresponding grounding of $Q$. This is because the truth value of groundings for which the part corresponding to $E$ is $false$ can never change to $true$, regardless of the assignments made to the part corresponding to $Q$, and thus do not affect the values assigned to the ground literals corresponding to $Q$ and are safely ignored by the inference. In other words,  because groundings only affect inference if their $E$ part is satisfied, $E$ can be viewed as ``selecting'' those groundings.

This view of $F$ allows us to specify two criteria for its effectiveness:  (1) among the groundings of $Q$ selected by $E$, how {\it uniformly} do we observe the pattern that $Q$ enforces in the data; and (2) how {\it surprising} is that pattern, i.e., how likely is one to observe the pattern in a randomly selected set of groundings of $Q$. Intuitively, the uniformity criterion measures the correctness of $F$. However, in the case of SRL models, we are equally interested in very correct formulas and very {\it in}correct ones, the latter getting negative weights during parameter training. The motivation for the second criterion is that we would like to find relational features that capture aspects beyond those that can be captured by simply using a prior over truth values. 

Next, we make these criteria precise. Let \set{L} be a set of ungrounded literals and let $G_{\set{L}}$ be a randomly chosen grounding of \set{L}. The joint assignment of truth values to the grounded literals in $G_{\set{L}}$ is a random variable $X_{\set{L}}$ with $2^{|\set{L}|}$ possible outcomes, i.e. if $\set{L}$ contains a single literal, the possible outcomes are $\{T, F\}$; if it contains two literals, the possible outcomes are $\{TT, TF, FT, FF\}$, etc.\footnote{Here, for simplicity we are ignoring the case where, after grounding \set{L}, some of the literals become identical.}  We are interested in the probability distribution that governs $X_{\set{L}}$. This distribution can be estimated empirically given observed truth assignments for a set of groundings $\set{I}_{\set{L}}$ of \set{L} by simple counting as the proportion of time a particular configuration of values is observed. \\
{\bf Definition 1.} Let $\mathbb{P}^{\set{I}}_{\set{L}}$ represent the empirical distribution over joint truth assignments to a randomly chosen grounding of \set{L}, as estimated on a set of groundings \set{I}.

Armed with this notation, we now go back to the view of a formula $F$ as consisting of a selector $E$ and an enforcer $Q$ and consider the empirical distribution $\mathbb{P}^{\set{I}_E}_{\set{Q}}$, where \set{Q} is the set of literals in $Q$, and $\set{I}_E$ is the set of groundings of \set{Q} selected by $E$, which is, in general, a subset of all possible groundings of \set{Q}. The first, uniformity, criterion identified above states that in an effective formula, $\mathbb{P}^{\set{I}_E}_{\set{Q}} (Q)$, the probability, according to $\mathbb{P}^{\set{I}_E}_{\set{Q}}$, of observing the configuration of truth values enforced by $Q$ should be as extreme as possible. \\
{\bf Criterion 1.} Effective formulas maximize $\max(\mathbb{P}^{\set{I}_E}_{\set{Q}} (Q), 1 - \mathbb{P}^{\set{I}_E}_{\set{Q}}(Q))$.

The second criterion states that $\mathbb{P}^{\set{I}_E}_{\set{Q}}$ should be significantly different from the ``default'' $\mathbb{P}^{\set{I}_{All}}_{\set{Q}}$, where $\set{I}_{All}$ is the set of all possible groundings of \set{Q}. In other words, we are interested in formulas whose selector $E$ homes in on groundings of \set{Q} for which the distribution over observed truth values deviates significantly from the default distribution over all possible groundings of \set{Q}.\\
{\bf Criterion 2.} Effective formulas maximize $dist(\mathbb{P}^{\set{I}_E}_{\set{Q}}, \mathbb{P}^{\set{I}_{All}}_{\set{Q}})$. 

Criterion 2 can be evaluated using standard measures of the distance between two distributions, such as KL divergence \cite{bishop:book06}, or by methods that are specifically designed to detect significant deviations on a large scale, e.g., \cite{dunning:jcl93}. 
\subsection{Simplifying Assumption}
However, evaluating such measures may be expensive. In particular, to estimate $\mathbb{P}^{\set{I}_{All}}_{\set{Q}}$, we need to enumerate the observed joint truth values for all possible $n^k$ groundings of \set{Q}, where $n$ is the number of entities in the domain and $k$ is the number of distinct variables appearing in \set{Q}. In general, this number is much higher than the number of groundings selected by $E$.  Instead, we note that relational domains are typically very sparse, i.e. the number of relations actually observed to be true is typically much smaller than the total number of possible relations that can form. Thus, rather than estimating $\mathbb{P}^{\set{I}_{All}}_{\set{Q}}$ for different sets \set{Q} from the data, we can make the approximately correct assumption that $\mathbb{P}^{\set{I}_{All}}_{\set{Q}}$ will be skewed towards configurations that involve $false$ assignments to the literals. In essence, this means that we can assume the same skewed default distribution $\mathbb{P}^{\set{I}_{All}}_{i}$ for all sets \set{Q} that contain $i$ literals. This assumption allows us to significantly simply the evaluation of rules according to the two criteria. Next, we describe how this is done for sets \set{Q} of different sizes.

The simplest case is when $|\set{Q}| = i = 1$, i.e., where the formula $F$ contains a single literal $Q_1$ of a target predicate. Supposing that $Q_1$ is non-negated, $\mathbb{P}^{\set{I}_{All}}_{i}$ is a Bernoulli distribution, which, because of the skew towards $false$ assignments, has a very small probability of success. Thus, to satisfy criterion 2 and maximize $dist(\mathbb{P}^{\set{I}_E}_{\set{Q}}, \mathbb{P}^{\set{I}_{All}}_{\set{Q}})$, $\mathbb{P}^{\set{I}_E}_{\set{Q}}$ needs to have a large probability of success. Combined with criterion 1, we note that the only way both criteria may be satisfied is if $\mathbb{P}^{\set{I}_E}_{\set{Q}}(Q_1)$ is maximized. Thus, when $|\set{Q}| = 1$, maximizing both criteria is as simple as requiring that the rule correctly identifies regions that contain high proportions of true positives. The case when $Q_1$ is negated is symmetric.

The situation is a bit less straightforward when $|\set{Q}| = i = 2$. For ease of exposition, let us assume that $\set{Q}$ contains two non-negated literals $Q_1$ and $Q_2$. Now, according to the sparsity assumption, $\mathbb{P}^{\set{I}_{All}}_{i}$ is skewed towards truth assignments in which at least one of $Q_1$ or $Q_2$ is $false.$ Thus, in order to satisfy criterion 2, we need formulas for which $\mathbb{P}^{\set{I}_E}_{\set{Q}}$ places significant mass on the case where $Q_1$ and $Q_2$ are both $true$. Combined with criterion 1, this means that effective formulas are ones for which $\mathbb{P}^{\set{I}_E}_{\set{Q}}(Q_1 \wedge Q_2)$ is large. Due to the sparsity in relational domains, in practice formulas with such selectors $E$ are rare. A second way to address criterion 2 is to look for formulas in which the conditional probability of one of $Q_1$ or $Q_2$ being $true$, given that the other one is $true,$ is surprisingly high. In terms of criterion 1, this translates into formulas for which $\mathbb{P}^{\set{I}_E}_{\set{Q}}(Q_1 \Rightarrow Q_2)$ is high. Thus, when $|\set{Q}| = 2$, \resolve\  autonomously determines the precise formulation of the enforcer $Q$ from the non-negated literals in \set{Q} by evaluating $\mathbb{P}^{\set{I}_E}_{\set{Q}}(Q_1 \wedge Q_2)$, $\mathbb{P}^{\set{I}_E}_{\set{Q}}(Q_1 \Rightarrow Q_2)$, and $\mathbb{P}^{\set{I}_E}_{\set{Q}}(Q_2 \Rightarrow Q_1)$ and choosing ones that result in high values. We note that the formulation $Q_1 \Rightarrow Q_2$ does not contradict our assumption that $F$ is a conjunction.  $Q_1 \Rightarrow Q_2$ can be expressed in conjunctive form as $\neg (Q_1 \wedge \neg Q_2)$, thus $F = E \wedge \neg (Q_1 \wedge \neg Q_2)$. When all literals in $E$ are observed, the effect of $F$ is equivalent to that of a formula $F' = E \wedge Q_1 \wedge \neg Q_2$, for which a negative weight is learned, despite the fact that $F$ and $F'$ are not logically equivalent.  

To summarize, in the general case, when $\set{Q}$ consists of $l$ literals $Q_1, Q_2, \dots, Q_l$, \resolve\ evaluates  
\[\mathbb{P}^{\set{I}_E}_{\set{Q}}(Q_1 \wedge Q_2 \dots \wedge \dots Q_l),\] and for each $k \in [1, l]$ 
\[\mathbb{P}^{\set{I}_E}_{\set{Q}}(Q_k | \wedge_{i \in [1, l], i \neq k} Q_i = true)\] and selects formulations that have high probabilities. 

This process is summarized in Algorithm~\ref{algo:resolve}. Steps 6-8 and 15-19 are only evaluated if the formula has more than one literal of a target predicate. The algorithm reduces the evaluation of the candidate formulas to the collection of a few statistics for each formula that can be easily computed on a stream of examples. Moreover, by taking advantage of the simplifying assumption, the algorithm avoids having to estimate $\mathbb{P}^{\set{I}_{All}}_{\set{Q}}$.
\begin{algorithm}[t]
\caption{\resolve\ Algorithm}
\label{algo:resolve}
\begin{algorithmic}[1]
\INPUT
\STATEA {\set{F}: set of formulas}\\
\STATEA {$\set{P}_T$: set of target predicates}\\
\STATEA{$\set{P}_E$: set of evidence predicates}\\
\STATEA{\set{S}: stream of training subgraphs}\\
\STATEA{$k_2$: number of  streamed training subgraphs to use}\\
\STATEA{$\theta$: threshold} \\
\OUTPUT
\STATEA{$\set{F}^*$: selected formulas}
\FOREACH{of the next $k_2$ subgraphs $s \in \set{S}$:}
   \FOREACH {$F \in \set{F}$}
      \STATE $E = $ sub-formula of $F$ consisting of literals with predicates in $\set{P}_E$
      \STATE $\set{Q} = \{Q_1, \dots, Q_l\}$ set of literals of $F$ with predicates in  $\set{P}_T$
      \STATE Compute $\mathbb{P}^{\set{I}_E}_{\set{Q}}(Q_1 \wedge Q_2 \dots \wedge \dots Q_l)$
      \FOREACH {$k \in [1, l]$}
         \STATE Compute $\mathbb{P}^{\set{I}_E}_{\set{Q}}(Q_k | \wedge_{i \in [1, l], i \neq k} Q_i)$
      \ENDFOR
   \ENDFOR
\ENDFOR
\FOREACH {$F \in \set{F}$}
  \IF { $Average(\mathbb{P}^{\set{I}_E}_{\set{Q}}(Q_1 \wedge Q_2 \dots \wedge \dots Q_l)) > \theta$}
    \STATE Add $E \wedge Q_1 \wedge Q_2 \dots \wedge \dots Q_l$ to $\set{F}^*$
  \ENDIF
 \FOREACH  {$k \in [1, l]$}
   \IF {$Average(\mathbb{P}^{\set{I}_E}_{\set{Q}}(Q_k | \wedge_{i \in [1, l], i \neq k} Q_i)) > \theta$}
     \STATE Add $E \wedge (Q_1 \wedge \dots\wedge Q_{k-1} \wedge Q_{k+1} \wedge \dots \wedge Q_l \Rightarrow Q_k)$ to $\set{F}^*$.
    \ENDIF
 \ENDFOR 
\ENDFOR
\end{algorithmic}
\end{algorithm}
\section{Experimental Evaluation}
\label{sect:experiments}
In this section, we demonstrate the methodology proposed in Section~\ref{sect:resolve} by developing applications in two social media domains. We compare \resolve\ to a system (called \baseline) that skips the second step outlined at the beginning of Section~\ref{sect:resolve} and directly trains weights on the formulas in the original set \set{F}. Because for formulas with $|\set{Q}| > 1$ \resolve\ automatically determines the correct logical connectives and negations in the part of the formula that involves  literals of the target predicate (i.e., lines 6-8 and 15-19 in Algorithm~\ref{algo:resolve}), the set \set{F} over which weights were trained by \baseline\ included all possible formulas considered by \resolve. The goal of our experiments was 1) to determine whether more accurate models are obtained with \resolve; and 2) to evaluate the relative efficiency of \resolve\ and \baseline. 

We implemented \resolve\ as part of the Alchemy system \cite{kok:tr05}. For weight-training, we adapted  Contrastive Divergence \cite{lowd:pkdd07} to learn from relational instances that arrive in a stream and used a Gaussian penalty on the weights. We preferred this algorithm over other available methods because we are interested in an efficient, light-weight approach. For inference, we used the MC-SAT algorithm \cite{poon:aaai06}. Generation of formulas from provided templates was implemented as a separate module in python.
\subsection{Data Sets}
The experiments were conducted in two social media domains -- WikiCollabs and Delicious. 
\subsubsection{WikiCollabs}
 The task in this domain is to predict project-specific collaborations in Wikipedia.\footnote{\url{http://en.wikipedia.org/wiki/Main_Page}} The data consists of all 3,538 Wikipedia articles that appeared in the featured\footnote{\url{http://en.wikipedia.org/wiki/Wikipedia:Featured_lists}} and controversial\footnote{\url{http://en.wikipedia.org/wiki/Wikipedia:List_of_controversial_issues}} lists in the period Oct. 7-21, 2009.  These articles are interesting because they are richly connected, both by their hyperlinks and by their human network of editors \cite{brandes:www09}. For each article, we collected the editors who contributed to it, either by directly editing the article, or by editing its ``Talk,'' i.e., discussion, page. Only edits that were not marked as ``minor'' by the editor were considered. In this way, we obtained a set of 280,068 editors.  In addition, we collected the hyperlinks among the articles. These articles are densely inter-linked, as indicated by the large number of hyperlinks (45,006) among them. 
Wikipedia articles often refer to external resources on the Web. Thus, for each article, we looked up the categorizations of each of its external references in the DMOZ open directory\footnote{\url{http://www.dmoz.org/}}. Because this information is not available for all URLs, we considered both exact matches of URLs, for which there were about 0.9 per article, as well as exact matches for just the domain name part of the URL, for which there were about 77 per article. 
An editor $E_1$ on Wikipedia can communicate with another editor $E_2$ by editing $E_2$'s ``Talk'' page. There were a total of 7,874,985 instances of communication between pairs of editors.
The set of evidence predicates $\set{P}_E$ included {\tt articleEdit(article, user)} and {\tt articleTalk(article, user)} for the two ways in which a user may contribute to an article; {\tt userTalk(user, user)}; {\tt hyperLink(article, article)}; {\tt similar(article, article)}, indicating that the cosine similarity between the tf/idf-weighted bag-of-words representation of the stemmed text in the two articles is between 0.1 and 0.5; {\tt verySimilar(article, article)}, indicating the similarity is greater than 0.5; {\tt category(article, category)}, providing the category under which the article appeared on the featured or controversial list, {\tt levelNExact(article, externalCategory)} and {\tt levelNInexact(article, externalCategory)} for the different levels $N = 1, 2, 3$ in the DMOZ hierarchy in which an external URL from an article is filed, either for exact URL matches or for matches of just the domain name of the URL.

The data from the WikiCollabs network was streamed in subgraphs $G_C$ that were centered around one of the editors $C$. $G_C$ contains all articles $A_C$ to which $C$ is related via the {\tt articleEdit} or {\tt articleTalk} predicates and all editors $E_C$, in addition to $C$ that contributed to any of the articles in $A_C$, as well as the other articles to which the editors in $E_C$ contributed. Also  included were any hyperlinks among the included articles, any instances of {\tt userTalk} relationships among the editors in $E_C$, any available category information on the included articles. The task was to learn to predict which editor in $E_C$ contributes to the articles in $A_C$, given all other information. For convenience, we represented the relationship between articles in $A_C$ and the users in $E_C$ by the {\tt modifies(article, user)} predicate, which was the only target predicate in $\set{P}_T$. Subgraphs were formed for users $C$ who made edits to the encyclopedia on at least 30 distinct days, had at least 30 collaborators, and edited at most 15 different articles. These restrictions are motivated by the observation that collaborator suggestion is most needed by editors who are strongly engaged with the encyclopedia, and so contribute to it over extended periods, but at the same time are focused in their interests. In this way, we excluded users, such as the ``60\% of registered users [who] never make another edit after their first 24 hours'' \cite{panciera:group09}, as well as users who help oversee the editing process and are therefore somewhat superficially involved in large numbers of edits, from having subgraphs $G_C$ formed around them. However, such users can still appear in the subgraph of another user. We obtained a total of 1785 subgraphs.

Formulas for the WikiCollabs task were generated from the templates shown in Table~\ref{wikiCollabsTemplates}. Predicates in all-caps are templates that get expanded in designer-specified ways, as shown in Table~\ref{wikiCollabsExpansions}.
\begin{table}[t]
\caption{Templates used to generate formulas in WikiCollabs.}
\label{wikiCollabsTemplates}
\begin{small}
\begin{align}
&EDIT(t1, u) \wedge SIMPLE\_REL(t1, t2) \Rightarrow \mathtt{modifies}(t2, u) \\
&EDIT(t1, u) \wedge LONG\_REL(t1, t2) \Rightarrow \mathtt{modifies}(t2, u) \\
&USER\_REL(u1, u2) \wedge \mathtt{modifies}(t, U1) \Rightarrow \mathtt{modifies}(t, u2)  \label{len2.1} \\
&USER\_REL(u1, u2) \wedge \mathtt{modifies}(t, U1) \wedge \mathtt{modifies}(t, u2)  \label{len2.2}
\end{align}
\end{small}
\end{table}
\begin{table}[t]
\caption{Expansions for the predicate templates used in WikiCollabs.}
\label{wikiCollabsExpansions}
\begin{small}
\begin{align*}
EDIT(t1, u) = &\{ \mathtt{articleEdit}(t1, u) | \mathtt{articleTalk}(t1, u)\}\\
SIMPLE\_REL(t1, t2) = &  \{ \mathtt{similar}(t1, t2) | \mathtt{verySimilar}(t1, t2) | \\
&\mathtt{hyperlink}(t1, t2) |\mathtt{hyperlink}(t2, t1) | \\
& \mathtt{category}(t1, c) \wedge \mathtt{category}(t2, c) \} \\
LONG\_REL(t1, t2) =& \{\mathtt{level[1|2|3]Exact}(t1, c) \wedge \mathtt{level[1|2|3]Exact}(t2, c) |\\
& \mathtt{level[1|2|3]Inexact}(t1, c) \wedge \mathtt{level[1|2|3]Inexact}(t2, c) \}\\
USER\_REL(u1, u2) = & \{\mathtt{userTalk}(u1, u2) | \\
& \mathtt{articleEdit}(t, u1) \wedge \mathtt{articleEdit}(t, u2)| \\
& \mathtt{articleTalk}(t, u1) \wedge \mathtt{articleTalk}(t, u2)|  \\
& \mathtt{articleEdit}(t, u1) \wedge \mathtt{articleTalk}(t, u2)| \\
& \mathtt{articleTalk}(t, u1) \wedge \mathtt{articleEdit}(t, u2)\}
\end{align*}
\end{small}
\end{table}
As can be seen from these expansions, the $EDIT$ template captures the different ways in which an editor may be related to an article, the $SIMPLE\_REL$ and $LONG\_REL$ expand to the different ways in which two articles may be related, and $USER\_REL$ expands to the different ways in which two users may be related. The $EDIT$ and $SIMPLE\_REL$ predicate templates were declared to be {\it compounders}, which means that when they are expanded in a rule, they can be replaced by a conjunction of more than one of their possible expansions. For example, in rule 1 above, $EDIT(t1, u)$ may be expanded to $\mathtt{articleEdit}(t1, u)$, or $\mathtt{articleTalk}(t1, u)$, or $\mathtt{articleEdit}(t1, u) \wedge \mathtt{articleTalk}(t1, u)$. We limited the length of compoundings to at most 2.  \resolve\ received only one version of rules generated from templates~\ref{len2.1} and~\ref{len2.2}, as it determines the correct configuration of literals of target predicates automatically. 
\subsubsection{Delicious}
The task in this data set is to predict user friendships on the Delicious social bookmarking site\footnote{\url{http://www.delicious.com/}} \cite{zhou:aaai10}.  We used the data collected by the authors of \cite{zhou:aaai10}, which includes 425,486 instances of the ``fan'' relationship, which indicates that one user is a fan of another one, 446,879 instances of the ``network'' relationship, which is the inverse of ``fan'' (i.e., if A is a fan of B, then B is in A's network), and 48,809,570 instances of the ternary ``tagging'' relationship between a user, a tag, and a URL. Although the ``fan'' and ``network'' relationships are inverses of one another, the observations in the data were not complete. We completed them by treating them as a single ``friendship'' relationship.

To stream this data, we formed subgraphs $G_C$, each of which was centered at one of the users $C$. The task was to predict all friendships of $C$. Each $G_C$ included $C$'s actual friends $Fr_C$ as true positives, and, as true negatives, a sampling of users who are friends with users from $Fr_C$. We did not form subgraphs for users $C$ for which the number of true negative friends was not at least as large as the number of true positive friends. The friendships between $C$ and the other users were hidden at test time, and the goal was to predict their existence. However, friendships among users other than $C$ were observed. For convenience in distinguishing between these two cases, we included an observed and an unobserved version of the {\tt friendship} relationship: an unobserved, i.e., target, {\tt cFriends(user)} predicate indicating that the user is friends with the implicit $C$, and an observed, i.e., evidence, {\tt friends(user, user)} predicate indicating that the two users are friends. For all users in $G_C$, we included observations about all URLs they bookmarked, along with the tags used. Those were captured via the following predicates: {\tt bkMark(page, user, tag)}; {\tt bkMarkAfter(page, user)}, which indicates that the user bookmarked the page at least one day after it was bookmarked by $C$; {\tt bkMarkBefore(page, user)}, {\tt bkMarkSameDay(page, user)}, which provide analogous information for pages bookmarked before or on the same day as bookmarked by $C$; {\tt usedTag(tag, user)}; {\tt sameTag(user, user)} and {\tt sameUrl(user, user)} to indicate, respectively, that two users (different from $C$) used the same tags and bookmarked the same URLs.
We used 656 subgraphs constructed in this way.

Formulas for the Delicious task were generated using the templates shown in Table~\ref{deliciousTemplates}. We used the expansions shown in Table~\ref{deliciousExpansions}.
\begin{table}[t]
\caption{Templates used to generate formulas in Delicious.}
\label{deliciousTemplates}
\begin{small}
\begin{align}
&REL(u1) \Rightarrow \mathtt{cFriends}(u1)\\
&LONG\_REL(u1) \Rightarrow \mathtt{cFriends}(u1) \\
&UREL(u1, u2) \wedge REL(u1) \Rightarrow \mathtt{cFriends}(u2)\\
&UREL(u1, u2) \wedge LONG\_REL(u1) \Rightarrow \mathtt{cFriends}(u2)\\
&UREL(u1, u2) \wedge \mathtt{cFriends}(u1) \Rightarrow \mathtt{cFriends}(u2) \label{len2.3}\\
&UREL(u1, u2) \wedge \mathtt{cFriends}(u1) \wedge \mathtt{cFriends}(u2) \label{len2.4}
\end{align} 
\end{small}
\end{table}
\begin{table}[t]
\caption{Expansions for the predicate templates used in Delicious}
\label{deliciousExpansions}
\begin{small}
\begin{align*}
REL(u1) = & \{ \mathtt{bkMarkAfter}(p, u1) | \mathtt{bkMarkBefore}(p, u1) |\mathtt{bkMarkSameDay}(p, u1) \}\\
LONG\_REL(u1) =& \{ \mathtt{usedTag}(t, u1) \wedge \mathtt{usedTag}(t, C) |  \mathtt{bkMark}(p, u1, t) \wedge \mathtt{bkMark}(p, C, t)\} \\
UREL(u1, u2) = &\{\mathtt{friends}(u1, u2) | \mathtt{sameTag}(u1,u2) | \mathtt{sameUrl}(u1, u2) \} 
\end{align*}
\end{small}
\end{table}
The $REL$ and $LONG\_REL$ templates expand to predicates that relate users to the user $C$ around whom the graph $G_C$ is centered via various bookmarking and tagging activities, whereas $UREL$ expands to different ways of relating two users.
$REL$ was declared a compounder, and $UREL$ was declared an {\it extender}, which meant that one or more possible expansions could be chained together. For example, $UREL(u1, u2)$ could be expanded in ways such as $\mathtt{friends}(u1, z1) \wedge \mathtt{sameUrl}(z1, u2)$. We allowed extensions and compoundings of length at most 2. As before, \resolve\ only needs the expansion from only one of the templates in lines~\ref{len2.3} and~\ref{len2.4}, as it determines the specific formulation autonomously.
\subsection{Methodology}
We performed four-fold cross-validation by splitting the subgraphs in the data randomly into 4 sets and performing 4 train/test runs, in each run withholding one of the folds for testing and training on the remaining three. We used $k_2 = 30$ and $\theta = 0.4$ in Algorithm~\ref{algo:resolve}. Before training weights, both for \resolve\ and \baseline,  we included a clause consisting of a single literal of the target predicate. This is standard practice in MLN applications that enables the model to capture the bias towards $false$ assignments by learning a negative weight on this single-literal clause.
 
The results are summarized using two standard metrics from the information retrieval literature \cite{manning:book08}:
\begin{itemize}
\item (MAP) \underline{M}ean \underline{a}verage \underline{p}recision, which is identical to the area under the precision-recall curve. The MAP score is computed over a set of test subgraphs $\mathcal{S}$ as follows:
\[ \textrm{MAP}(\mathcal{S}) = \frac{1}{|\mathcal{S}|} \sum_{s \in \mathcal{S}}\frac{1}{|R_s|} \sum_{r \in R_s} P@r. \]
Above, $R_s$ is the set of all possible $\mathtt{(p, c)}$ pairs, and the precision at $r$ is defined as  
\[P@r = \frac{\textrm{Num of true positive pairs among the top } r}{r} \]
\item (AUC-ROC) Area under the ROC Curve, which is identical to the mean average true negative rate. This score is computed as follows:
\[ \textrm{AUC-ROC}(\mathcal{S}) = \frac{1}{|\mathcal{S}|} \sum_{s \in \mathcal{S}} \frac{1}{|R_s|} \sum_{r \in R_s} TN@r,\]
where the true negative rate at $r$ is defined as
\[ TN@r = \frac{\textrm{Number of true negatives below position } r}{\textrm{Total num true negatives}}.\]
\end{itemize}
\subsection{Results}
\begin{figure}[t]
\begin{center}
\begin{tabular}{ccc}
{\bf WikiCollabs} & & {\bf Delicious} \\
\includegraphics[scale=0.45]{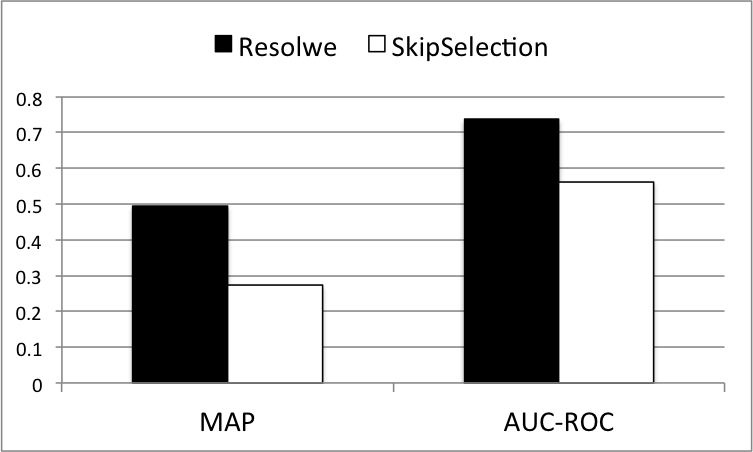}  & &
\includegraphics[scale=0.45]{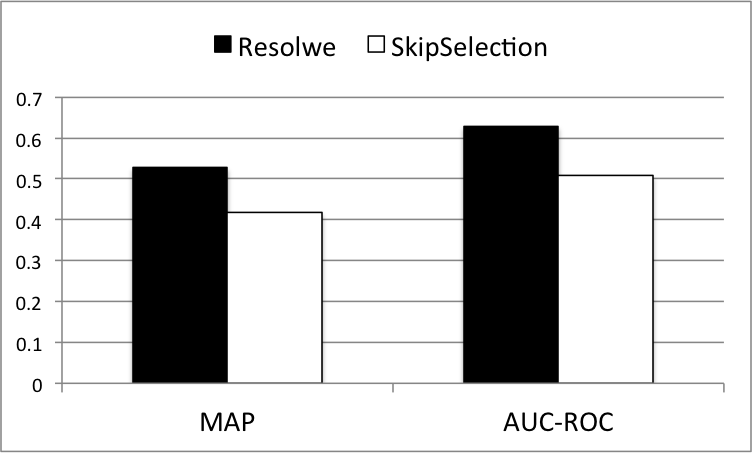} 
\end{tabular}
\end{center}
\caption{Comparison between \resolve\ and \baseline\ in terms of Mean Average Precision and Area under the ROC curve in WikiCollabs (left) and Delicious (right). Observed differences are significant at the 0.001 level.}
\label{fig:results}
\end{figure}

The results of our experiments are shown in Figure~\ref{fig:results}. All differences in this figure are significant at the 0.001 level according to a paired t-test. As can be seen, selecting formulas with \resolve\ before training weights leads to significant improvements in both domains according to both metrics. Because the AUC-ROC performance of a random predictor would be 0.5, we can see that, in fact, by using \resolve, we can go from near-random performance, to significantly higher accuracy levels. Moreover, \resolve\ learns significantly faster than \baseline. Table~\ref{tbl:timing} presents results for the number of minutes taken by \resolve\ and by weight learning on dedicated Xeon 2.67GHz CPUs, averaged over the 4 folds in each domain. In both cases, using \resolve\ leads to dramatic decrease in training time.

 We note that our results in the Delicious domain are not comparable to those of Zhou et al.~\cite{zhou:aaai10} because their system uses global computations over all available data to arrive at predictions, whereas here we focus on making predictions using information local to subgraphs of the original relational graph. 

\begin{table}[t]
\caption{Training time in minutes for completing steps 2 (formula selection) and 3 (parameter learning), as outlined at the beginning of Section~\ref{sect:resolve}, averaged over 4 folds.}
\label{tbl:timing}
\begin{center}
\begin{tabular}{|c||c|c|c||c|c|c|} \hline
& \multicolumn{3}{|c||}{{\bf WikiCollabs}} & \multicolumn{3}{c|}{{\bf Delicious}} \\ \hline
& Step 2& Step 3& Total & Step 2& Step 3& Total\\ \hline
\resolve &  94.01 & 91.15 & 185.16 &30.66 & 62.76 & 93.42\\
\baseline & -& 3236.40 & 3236.40 & - & 602.08& 602.08\\\hline
\end{tabular}
\end{center}
\end{table}

\section{Related Work}
Structure learning and feature selection are important problems that have been widely studied in both relational and i.i.d. settings. Most feature selection approaches, e.g., \cite{guyon:jmlr03}, have been developed for non-streaming classification settings. One recent exception is the work of Wu et al. \cite{wu:icml10}, who study a classification task where the features arrive in a stream, while the data set is fixed. In contrast, here we explore the setting where the pool of features is fixed, but the data arrives as a stream.

Closely related to this paper is work on structure learning of statistical relational models \cite{getoor:jmlr02,kok:icml05,mihalkova:icml07,biba:ilp08,huynh:icml08,kok:icml09,kok:icml10,khosravi:aaai10}. This literature has made important advances on focusing the search through the super-exponential space of candidate models, thus discovering more accurate candidates faster. Less emphasis has been placed on how to evaluate candidate structures and, in most existing work, evaluation has been carried out by computing a probabilistic score over candidate structures that, crucially, assumes that the training data is available in memory. In contrast, this paper addresses the complementary setting  common in many relational applications where sufficient background knowledge is available to generate candidate structures, and the challenge is in how to efficiently evaluate them on data that is presented to the learner in a stream. The set-up explored here is probably most similar to that assumed by Huynh and Mooney \cite{huynh:icml08}, where formula selection and parameter training are carried out in two separate stages. However, while that previous work also employs an accuracy-based measure (that of {\sc Aleph} \cite{srinivasan:01}) to evaluate rule candidates, it does not address the task of evaluating candidates that have more than a single literal of the target predicate and does not consider streaming the relational instances.

A few authors have addressed learning of structure from data streams. Dries and De Raedt \cite{dries:ilp09} introduced an inductive logic programming technique that uses candidate elimination to learn theories from a stream of examples. Their work applies to noise-free data. Recently, Kummerfeld and Danks \cite{kummerfeld:tr10} introduced a ``Temporal Difference Structure Learning'' Algorithm that learns causal structure from a data stream. This algorithm targets causal discovery in graphical models and is not applicable to the relational setting assumed here. 

Learning from data streams in relational settings has so far focused on training the parameters of a model for which structure is provided (as done in \baseline, described in Section~\ref{sect:experiments}). This approach was adopted by Mihalkova and Mooney \cite{mihalkova:ecml09} 
and in upcoming work by Huynh and Mooney \cite{huynh:sdm11}.  
\section{Conclusion}
We proposed an approach to streamlining application development in relational domains by efficiently evaluating a set of candidate formulas on relational instances that are streamed one at a time. The evaluation algorithm is derived from two natural criteria and efficiency is achieved by exploiting the fact that typical relational domains are sparse. We fleshed out our approach to develop two applications in large and noisy social media tasks, demonstrating significant gains in the speed and accuracy of learning. 

Avenues for future work include adapting this approach to tackle domains that experience gradual concept drift. One way to do this is to interleave steps 2 and 3 outlined at the beginning of Section~\ref{sect:resolve} and use a decaying average of the statistics calculated by Algorithm~\ref{algo:resolve}. As soon as step 2 determines a change in the structure over which weights are learned in step 3, the change is implemented, keeping the weights of the remaining rules at their currently learned values, and the process continues. A second potential direction for future work is exploiting other ways in which relational data may be streamed to the learner. For example, one interesting setting arises when the learner is allowed to actively decide in what ways and how much to grow the subgraphs $G_C$ around entities of interest $C$. 
\section{Acknowledgements}
We would like to thank Tom Chao Zhou for providing us with the Delicious data set.  L. Mihalkova is supported by a CI fellowship under NSF Grant \# 0937060 to the Computing Research Association.

\end{document}